\documentclass{article}
\usepackage[utf8]{inputenc}

\usepackage[preprint]{neurips_2026}

\usepackage[utf8]{inputenc} 
\usepackage[T1]{fontenc}    
\usepackage{hyperref}       
\usepackage{url}            
\usepackage{booktabs}       
\usepackage{amsfonts}       
\usepackage{nicefrac}       
\usepackage{microtype}      

\usepackage{graphicx}
\usepackage{cyknn}
\usepackage{amsmath}
\usepackage{subfig}
\usepackage{algorithm}
\usepackage[noend]{algpseudocode}
\usepackage{setspace}
\AtBeginEnvironment{algorithmic}{\setstretch{0.9}} 

\usepackage{comment}

\def\CYKmatrix#1{\mathbf{#1}}
\def\transpose{T}

\definecolor{cadmiumgreen}{rgb}{0.0, 0.42, 0.24}

\title{Neuro-symbolic Syntactic Parsing: Shaping a Neural Network with the CYK Algorithm}
\author{Fabio Massimo Zanzotto$^{\ast}$, Federico Ranaldi$^{\ast}$, Giorgio Satta$^{+}$\\
$^{\ast}$ Human-centric ART, University of Rome Tor Vergata, Italy\\
$^{+}$ DEI, University of Padua, Italy\\
}

\begin{document}

\maketitle

\begin{abstract}
In this paper, we show the possibility of a direct injection of algorithms into neural network architecture. We focus on a complex algorithm, that is, Cocke-Youger-Kasami (CYK) for parsing context-free grammars in Chomsky Normal Form and we propose CYKNN, a simple recurrent neural network architecture for encoding the CYK algorithm in trainable matrix-vector multiplications.We experimented with a very simple grammar with 4 variations showing that our approach outperforms existing LLMs with more than 20B parameters with an in-context learning setting and smaller LLMs of the Qwen family fine-tuned with LoRA. Our attempt paves the way to a different approach to neuro-symbolic methodologies. 
\end{abstract}

\section{Introduction}



 The ability of transformer-based architectures \citep{NIPS2017_3f5ee243} to \emph{grok} simple algorithmic capabilities from data is particularly fascinating, in both shallow \citep{power_grokking_2022} and  deeper transformer architectures \citep{DeepSeekAI2025DeepSeekR1IR}. The underlying mechanisms driving this phenomenon remain a subject of active debate and are being examined from multiple theoretical and empirical perspectives \citep{nanda_progress_2023,tan_understanding_2024,huang_unified_2024}. Moreover, it appears that these capabilities are acquired within a single training epoch in large language models \citep{li_grokking_2026}, whereas \emph{grokking} in specialized shallow transformer architectures typically emerges only after multiple training epochs, subsequent to an initial phase of overfitting on the training data. 

Demanding transformer-based architectures to extrapolate algorithmic knowledge from training data is an exaggerated expectation that exceeds what is required to human learners and, in fact, transformer-based Large Language Models do not succeed in  learning complex algorithms \citep{mirzadeh_gsm-symbolic_2025,shojaee_illusion_2025}. Indeed, this request calls for the most challenging form of generalization. Extrapolating algorithmic knowledge requires years to human experts and the fact that specific algorithms already exist does not facilitate the task for transformers that have to discover it from training data. Some reserach suggests that these transformers are only large language memories \citep{zanzotto-etal-2025-position}.



For decades, programmers have instructed machines with algorithms to resolve specific tasks and machine learning was confined to dealing with cases that cannot be handled algorithmically where answers should be found also for possibly unexpected data. These days, machine learning specialists are asking to neural networks to learn what can be programmed. Although understanding this generalization ability of transformers is fascinating, a tantalizing questions arises: if algorithms are known, is there a way to encode directly these algorithms in a neural network like architecture governed by matrix-vector multiplication?

Coding algorithms directly in neural networks is a possible, intriguing alternative to the current trend of passively relying on emerging capabilities of transformers. Indeed, this idea stems from a debate dating back to the late '80s: 
do neural networks with their way of representing information have extra capabilities with respect to classical symbolic approaches underlying the activity of programming machines? In case of a negative answer, the alternative is viable. \citet{FODOR19883} suggest that distributed representations underlying neural network architectures are ``only an implementation of the Classical approach” where classical approach is related to discrete symbolic representations. \citet{Chalmers1992} affirms that distributed representations give the important opportunity to 
reason \textit{holistically} about encoded knowledge. This means that decisions over some specific part of the stored knowledge can be taken without retrieving the specific part but acting on the whole representation. Substantially, neural networks and symbolic approaches are somehow similar even if large language models realized with transformer-based neural network architectures are astonishing.

In this paper, we pave the way for the direct injection of algorithms into neural network architecture by showing that we can encode at least a complex algorithm, that is, Cocke-Younger-Kasami (CYK) for parsing context-free grammars in Chomsky Normal Form. We stem from the initial studies of \citep{a13100262} and we propose CYKNN, a simple recurrent neural network architecture for encoding the Cocke-Younger-Kasami (CYK) in trainable matrix-vector multiplications. In some sense, the injection of the algorithm shapes the architecture of the CYKNN. Stemming from a way to represent data treated by the algorithm, the algorithmic operations are translated into matrix-vector multiplications, which \textit{holistically} change the status of the data. More concretely, holistically means changing \textit{for}-cycles in a single matrix-vector operation. We experimented with a very simple grammar with 4 variations showing that our approach outperforms existing LLMs with more than 20B parameters with an in-context learning setting \citep{openai2025gptoss120bgptoss20bmodel,gemmateam2025gemma3technicalreport} and smaller LLMs of the Qwen family \citep{qwen2025qwen25technicalreport,yang2025qwen3technicalreport} fine-tuned with LoRA \citep{10.1145/3676151.3719377}.

\section{Background and Notation}
\label{sec:preliminaries}

In this~section, we introduce the~basics about the~CYK algorithm and~overview a class of distributed representations called holographic reduced representation. 

\subsection{Why CKY parsing with context-free-grammars is important and what is the CYK algorithm?}

Parsing algorithms based on context-free grammars (CFGs) have been central to computer science since the 1960s: these algorithms are the base for parsing programming languages as well as for parsing natural language. Hence, parsing algorithms fall into two broad classes, depending on whether or not they handle ambiguous grammars. For unambiguous grammars generally used for programming language compilers, specialized techniques such as LL and LR~\citep{Sippu:90} parse input efficiently in \textit{linear time}~\citep{Aho-et-al:2006}. For ambiguous grammars, general parsing methods construct compact parse forests representing all parse trees accepted by the grammar for a given input string. These are generally dynamic-programming algorithms that run in \textit{polynomial time} in the input length (see~\citep{Graham:76} for an overview). Parsing algorithms for ambiguous CFGs are widely used in areas such as natural language processing, where different parse trees correspond to different semantic interpretations~\citep{Nederhof-Satta:2010}.

The~Cocke--Younger--Kasami algorithm (CYK) 
\citep{Cocke:1969,Younger1967,Kasami:1965:CYK} was 
the~first parsing method for ambiguous CFGs to be discovered. 
This~algorithm is at the~core of many common algorithms 
for natural language parsing, where it is used in combination 
with probabilistic methods and~other machine learning techniques
to estimate the probabilities for grammar rules based on large datasets%
~\citep{Charniak:1996}, as well as to produce 
a parse forest of all parse trees for an input string 
and~retrieve the~trees having the overall 
highest probability~\citep{huang-chiang-2005-better}. 

The CYK algorithm is a classical dynamic-programming algorithm for parsing and recognition based on context-free grammars (CFGs). We provide here a brief description of the algorithm in order to introduce the notation used in later sections; we closely follow the presentation in~\citep{Graham:76}, and we assume the reader is familiar with the formalism of CFG~(\citep{Hopcroft:01}, Chapter~5). The algorithm requires CFGs in Chomsky normal form, where each rule has the form $A \rightarrow B C$ or $A \rightarrow a$, where $A$, $B$, and $C$ are nonterminal symbols and $ a$ is a terminal symbol. We write $R$ to denote the set of all rules of the grammar and $ NT$ to denote the set of all its nonterminals. 

Given an input string $w = a_1 a_2 \cdots a_n$, $n \geq 1$, where each $a_i$ is an alphabet symbol, the~algorithm uses a two-dimensional table $P$ of size $(n+1) \times (n+1)$, where each entry stores a set of nonterminals representing partial parses of the~input string. More precisely, for $0 \leq i < j \leq n$, a nonterminal $A$ belongs to set $P[i,j]$ if and only if there exists a parse tree with root $A$ generating the substring $a_{i+1} \cdots a_j$ of $w$. Thus, $w$ can be parsed if the~initial nonterminal of the~grammar $S$ is added to $P[0,n]$. 
Algorithm~\ref{alg:CYK} shows how table $P$ is populated. 
$P$ is first initialized using unary rules, at Line~\ref{l:init}. 
Then,~each entry $P[i,j]$ is filled at Line~\ref{l:binary} 
by looking at pairs $P[i,k]$ and~$P[k,j]$ and~by using binary rules. 
\begin{figure}[H]
\begin{center}
{\centering
\begin{minipage}{.55\linewidth}
\begin{algorithm}[H]
\setlength\baselineskip{18.5pt}
\begin{algorithmic}[1]
\For {$i \gets 1$ to $n$}
 \For {each $A \rightarrow a_i$ in $R$}
 	\State add $A$ to $P[i-1,i]$ \label{l:init}
 \EndFor
\EndFor
\For {$j \gets 2$ to $n$}
	\For {$i \gets j-2 $ to $0$}
		\For {$k \gets i+1$ to $j-1$}
 		\For {each $A \rightarrow B C$ in $R$}
				\If {$B \in P[i,k]$ and~$C \in P[k,j]$} 
					\State add $A$ to $P[i,j]$ \label{l:binary}
				\EndIf 
 		\EndFor
 	\EndFor
 \EndFor
\EndFor
\end{algorithmic}
\caption{CYK(string $w = a_1 a_2 \cdots a_n$, rule set $R$)\\ \Return table $P$.}\label{alg:CYK}
\end{algorithm}
\end{minipage}
}
\begin{tabular}{cc}
Grammar rules $R$ & Table $P$ \\
\begin{tabular}{c}
$\begin{array}{ccc}
S &\rightarrow & D E\\
S &\rightarrow & D S\\
D &\rightarrow & a\\
E &\rightarrow & b\\
\end{array}$
\end{tabular}
&
\begin{tabular}{c}
\setlength{\tabcolsep}{0.00001em}
\begin{tabular}{ccc}

\cline{1-3} 
\cykcell{(0,1)}{D} & \cykcell{(0,2)}{} & \cykcell{(0,3)}{S} \\ 
\cline{1-3} 
 & \cykcell{(1,2)}{D} & \cykcell{(1,3)}{S} \\ 
\cline{2-3} 
 & & \cykcell{(2,3)}{E} \\ 
\cline{3-3} 
\end{tabular}
\end{tabular}
\end{tabular}
\caption{The Cocke--Younger--Kasami (CYK) parsing algoritm, a simple context-free grammar (CFG), the CYK parsing table $P$ 
for the~input string $w = aab$}
\label{fig:cyk_matrix}
\end{center}
\end{figure}

A running example is presented in Figure~\ref{fig:cyk_matrix}, showing a set $R$ of grammar rules along with the~table $P$ produced by the~algorithm when processing the~input string $w = aab$.
For instance, $S$ is added to $P[1,3]$, since $D \in P[1,2]$, $E \in P[2,3]$, and~$(S \rightarrow D E) \in R$. Since $S \in P[0,3]$, we conclude that $w$ can be generated by the grammar. 

\subsection{Holographic Reduced Representations to Encode Symbolic Reasoning in Neural Networks}
\label{sec:revisedhrr}

Holographic Reduced Representations (HRRs), originally proposed in~\citep{Plate1995} and later extended in~\citep{Zanzotto:ICML:2012}, are distributed representations that are particularly suitable for our goal of encoding the two-dimensional CYK parsing table $P$ and for implementing the operation that retrieves the content of its cells $P[i,j]$. 
In what follows, we present the operations we employ, together with a graphical notation to illustrate their properties based on Tetris-like pieces.

In line with \cite{a13100262}, we use a revised version of extended HRRs that 
exploits the shuffled circular convolution in the description of the symbols. 
Each symbol $a$ is represented as a matrix 
$\mysc{a} = \cm{A}\Phi$ in $\R^{d\times d}$ for some constant $d$, where $\cm{A}$ is the circulant matrix of $a$ and $\Phi$ is a permutation matrix. 
Basically, given a symbol $a$, $\mysc{a}$ is the matrix representing its random vector $\vec{a}$ ready for the shuffled circular convolution. Along with $\mysc{a}$, it is possible to define the inverse representation $\invsc{b} =\Phi^\transpose\cm{A}^\transpose$ of a symbol $b$. The matrix multiplication $\mysc{a} \invsc{b}$ has then the following property:
\begin{eqnarray*}
\mysc{a} \invsc{b} & \approx & \begin{cases}
\CYKmatrix{I} & \text{if } \vec{a} = \vec{b} \\
\CYKmatrix{0} & \text{if } \vec{a} \neq \vec{b} \\
\end{cases} 
\end{eqnarray*}
(see Appendix \ref{sec:properties} for details).

\begin{figure}[htbp]
\subfloat[Tetris-like visualization of representations of symbols]{%
\begin{tabular}{c}
$\mysc{a}$ = \symbA{}, $\invsc{a}$ = \rsymbA{}, \\
$\mysc{b}$ = \symbB{}, $\invsc{b}$ = \rsymbB{}
\end{tabular}
}\hfill
\subfloat[Visulazation of the sequence $abS$]{
\begin{tabular}{c}
 $\mysc{a} \mysc{b} \mysc{S}$\\
\symbA\symbB\symbS 
\end{tabular}
}\hfill
\subfloat[Application of an inverse symbol to a sequence]{%
\begin{tabular}{ccc}
$\invsc{a}\mysc{a}\mysc{b}\mysc{S}$ & $\approx$ & $\CYKmatrix{I}\mysc{b}\mysc{S} = \mysc{b}\mysc{S}$\\
\rsymbA \symbA\symbB\symbS &  & \symbB\symbS
\end{tabular}
}\\
\subfloat[Representation of the set of sequences $\{ abS, DSa \}$]{%
\begin{tabular}{ccc}
$\CYKmatrix{L}$ & = & $\mysc{a}\mysc{b}\mysc{S} + \mysc{D}\mysc{S}\mysc{a}$ \\
 & & \fbox{\symbA\symbB\symbS \hspace{0.5cm} \symbD\symbS\symbA}
\end{tabular}
} \hspace{2cm} 
\subfloat[Application of a symbol to the set 
of sequences $\CYKmatrix{L}$ resulting 
in set $\CYKmatrix{L}'$]{%
\begin{tabular}{c}
\begin{tabular}{ccccc}
$\CYKmatrix{L}'$ & = & \rsymbA$\CYKmatrix{L}$ & = & \fbox{\symbB\symbS }
\end{tabular}
\end{tabular}
}
\caption{Tetris-like representation of matrix representations of symbols}
\label{fig:tetris_reperesentation}
\end{figure}

The above representations have the interesting feature that they delete symbols in a chain of matrix multiplications whenever representations of symbols appear next to their inverse representations. In fact, any two adjacent 
inverse representations of symbols are replaced by the nearly 
identity matrix, which is invariant for matrix multiplication. 
As this behavior 
is reminiscent of what happens in Tetris, where pieces eliminate lines whenever their shapes perfectly fill the gaps, \citet{a13100262} adopted a Tetris-like visualization for the above operations (see Fig. \ref{fig:tetris_reperesentation}.(a)).
In this visualization, sequences of symbols correspond to sequences of pieces (see Fig. \ref{fig:tetris_reperesentation}.(b)).
Then, as in Tetris, elements whose shapes complement each other cancel out and are removed from the sequence (see Fig. \ref{fig:tetris_reperesentation}.(c)). Collections of sequences of symbols (sums of matrix products representing sequences) are depicted inside boxes (see Fig. \ref{fig:tetris_reperesentation}.(d)). In addition to the~usual Tetris rules, we assume here that an element with a certain shape selects from a box only elements with the~complementary shape facing it (see Fig. \ref{fig:tetris_reperesentation}.(e)).

Using  the~Tetris metaphor, we describe our new model to encode $P$ tables as matrices of real numbers $\CYKmatrix{P}$, and how we can implement CFG rule applications by means of matrix multiplication.

\section{Encoding the CYK algorithm in a Neural Network}

Our goal is to demonstrate that a symbolic algorithm can be expressed in terms of linear algebraic operations, that is, matrix-vector multiplication. By using this new formulation, we can systematically transform a symbolic algorithm into an equivalent formulation that induces a corresponding neural network architecture. To demonstrate this possibility, we will use the well-known CYK algorithm.

In the following, we first describe the inspiring idea and then the corresponding neural network, the CYKNN, designed to perform the CYK algorithm steps. 

\subsection{The Inspiring Idea}
With respect to what is proposed in \citep{a13100262}, our methodology to build a neural network shaped by a CYK algorithm stems from a simple, game-changing idea: associating the representation of the cells of CKY table with a convenient way of representing rules so that the update operation of the table is simple and can be done holistically. Let's represent symbols in cells $A \in P[i,j]$ in a distribued matrix $\CYKmatrix{P}$ as: 
$$
   \mysc{i} \mysc{A} \mysc{j}
$$
In this case, the overall matrix $\CYKmatrix{P}$ for the running example in Figure \ref{fig:cyk_matrix} is: 
\begin{center}
\begin{tabular}{c}
\fbox{\parbox{\dimexpr\linewidth-75\fboxsep-2\fboxrule\relax}{\centering 
\begin{tabular}{ccc}
\indexZero  \symbS \indexThree
&
\indexOne  \symbS \indexThree
\end{tabular}\\
\begin{tabular}{ccc}
\indexZero\symbD \indexOne
&
\indexOne\symbD \indexTwo
&
\indexTwo\symbE \indexThree
\end{tabular}

}}
\end{tabular}
\end{center}

If this representation is associated with a correlated 
formalization of grammar rules, the update operation may become very simple. Binary grammar rules $A \rightarrow B C$ can be represented in the following way: 
$$
\CYKmatrix{W} = \sum_{A \rightarrow B C \in R} \invsc{B} \mysc{A} \invsc{C} 
$$
Then the matrix representation $\CYKmatrix{R}$ of set $R$ is the following: 
$$
\CYKmatrix{R} = \sum_{k=1}^n \invsc{k}\CYKmatrix{W}\invsc{k}
$$

\begin{center}
\begin{minipage}{.55\linewidth}
\begin{algorithm}[H]
\setlength\baselineskip{18.5pt}
\begin{algorithmic}[1]
\State $\CYKmatrix{P} \gets \CYKmatrix{P}_0$ 
\For {$i \gets 0$ to $n-1$}
   \State $\CYKmatrix{P}' \gets \CYKmatrix{P}\CYKmatrix{R}\CYKmatrix{P} $
   \State $\CYKmatrix{P} \gets \CYKmatrix{P} + \CYKmatrix{P}'$ 
\EndFor
\end{algorithmic}
\caption{ProtoCKYNN($\CYKmatrix{P}_0$) \Return table $P$}
\label{alg:protoCYKNN}
\end{algorithm}
\end{minipage}
\end{center}

The matrix $\CYKmatrix{R}$ along with the matrix $\CYKmatrix{P}$, which represents the CYK matrix, can be used to determine the next step starting from a previous status. If the matrix $\CYKmatrix{P}$ is filled with all the constituents up to a given length, the product $\CYKmatrix{P}' = \CYKmatrix{P} \CYKmatrix{R} \CYKmatrix{P}$ will operate on addends that can result in the following way. For each $i,k,j$  and for every rule$A \rightarrow B C$ there will be addends of this kind:
$$
   \mysc{i} \mysc{B} \mysc{k} \invsc{k} \invsc{B} \mysc{A} \invsc{C} \invsc{k} \mysc{k} \mysc{C} \mysc{j}  =  \mysc{i} \mysc{A} \mysc{j}
$$
that will add to $\CYKmatrix{P}'$ constituents 
of length $j-i$. 
The CKY parser will then be simplified into the cycle described in Algorithm \ref{alg:protoCYKNN}.
At the end of the cycle, the matrix $\CYKmatrix{P}$ will contain all the constituents related to the input sentence. 
The proposed algorithm is overly simplified and, as such, cannot be expected to be fully accurate. In particular, the reduction in computational complexity from \(O(n^3)\) to \(O(n)\) is achieved at the cost of introducing an approximation at each iteration of the processing cycle. To mitigate this limitation, in the neural network implementation, we incorporated a signal amplification and refinement mechanism for the representations of constituent symbols, which is applied at every step of the procedure.

To clarify the underlying concept, we employ our ongoing example in Fig. \ref{fig:cyk_matrix}. We begin with a CYK matrix that contains only the first diagonal, namely
\(\mathit{CYK}_{\text{matrix}} = \{(0,1,D), (1,2,D), (2,3,E)\}\).
In the distributed representation, this matrix is expressed as follows:
$$
\CYKmatrix{P} = \fbox{\indexZero\symbD \indexOne \hspace{0.1cm} \indexOne\symbD \indexTwo \hspace{0.1cm} \indexTwo\symbE \indexThree }
$$
In addition to the status of the matrix $\CYKmatrix{P}$, a simplified rule matrix $\CYKmatrix{R}$ representing only the binary rule  $S \rightarrow DE$ 
is given by:
$$
\CYKmatrix{R} = \fbox{\rindexOne\rsymbD\symbS\rsymbE\rindexOne \hspace{0.1cm} \rindexTwo\rsymbD\symbS\rsymbE\rindexTwo}
$$
The application of the principal operation \(P' = PRP\) within the iterative cycle of Algorithm~\ref{alg:protoCYKNN} yields the following effect:
$$ 
\CYKmatrix{P}' = \CYKmatrix{P}\CYKmatrix{R}\CYKmatrix{P} = \fbox{\indexZero\symbD \indexOne \hspace{0.1cm} \indexOne\symbD \indexTwo \hspace{0.1cm} \indexTwo\symbE \indexThree } \;\fbox{\rindexOne\rsymbD\symbS\rsymbE\rindexOne \hspace{0.1cm} \rindexTwo\rsymbD\symbS\rsymbE\rindexTwo} \; \CYKmatrix{P}  = 
$$
Focusing on the first $\CYKmatrix{P}\CYKmatrix{R}$, the matrix multiplication creates all the following sums of matrices:
$$ 
= \fbox{\indexZero\symbD \indexOne\rindexOne\rsymbD\symbS\rsymbE\rindexOne \hspace{0.1cm} \indexOne\symbD \indexTwo \rindexOne\rsymbD\symbS\rsymbE\rindexOne \hspace{0.1cm} \indexTwo\symbE\indexThree \rindexOne\rsymbD\symbS\rsymbE\rindexOne \hspace{0.1cm}
\indexZero\symbD \indexOne \rindexTwo\rsymbD\symbS\rsymbE\rindexTwo \hspace{0.1cm} \indexOne\symbD \indexTwo \rindexTwo\rsymbD\symbS\rsymbE\rindexTwo \hspace{0.1cm} \indexTwo\symbE \indexThree \rindexTwo\rsymbD\symbS\rsymbE\rindexTwo}\; \CYKmatrix{P}  = 
$$
Some of these addends contain adjacent matrices and their inverses, which can be algebraically canceled:
$$ 
= \fbox{\indexZero\symbS\rsymbE\rindexOne \hspace{0.1cm} \indexOne\symbD \indexTwo \rindexOne\rsymbD\symbS\rsymbE\rindexOne \hspace{0.1cm} \indexTwo\symbE\indexThree \rindexOne\rsymbD\symbS\rsymbE\rindexOne \hspace{0.1cm} \indexZero\symbD \indexOne \rindexTwo\rsymbD\symbS\rsymbE\rindexTwo \hspace{0.1cm} \indexOne\symbS\rsymbE\rindexTwo \hspace{0.1cm} \indexTwo\symbE \indexThree \rindexTwo\rsymbD\symbS\rsymbE\rindexTwo}\; \CYKmatrix{P}  \approx 
$$
Upon removal of extended matrix sequences, identified as noise, from the visualization, the following behavior is observed:
$$ 
\approx \fbox{\indexZero\symbS\rsymbE\rindexOne \hspace{0.1cm} \indexOne\symbS\rsymbE\rindexTwo}\; \fbox{\indexZero\symbD \indexOne \hspace{0.1cm} \indexOne\symbD \indexTwo \hspace{0.1cm} \indexTwo\symbE \indexThree }  \approx \fbox{\indexOne\symbS\rsymbE\rindexTwo \indexTwo\symbE \indexThree} \approx \fbox{\indexOne\symbS\indexThree} 
$$
Then, finally, the application of this single operation generates the filling of the new diagonal, taking into account longer sequences, that is, $\CYKmatrix{P}'$ contains $(1,3,S)$.
This is a clear example of \textit{holistic reasoning}.

\subsection{CYK Neural Network (CYKNN)}

\begin{figure}[htbp]
\subfloat[Recurrent CYKNN Architecture]{%
\begin{minipage}[b]{.45\linewidth}
\includegraphics[width=\linewidth]{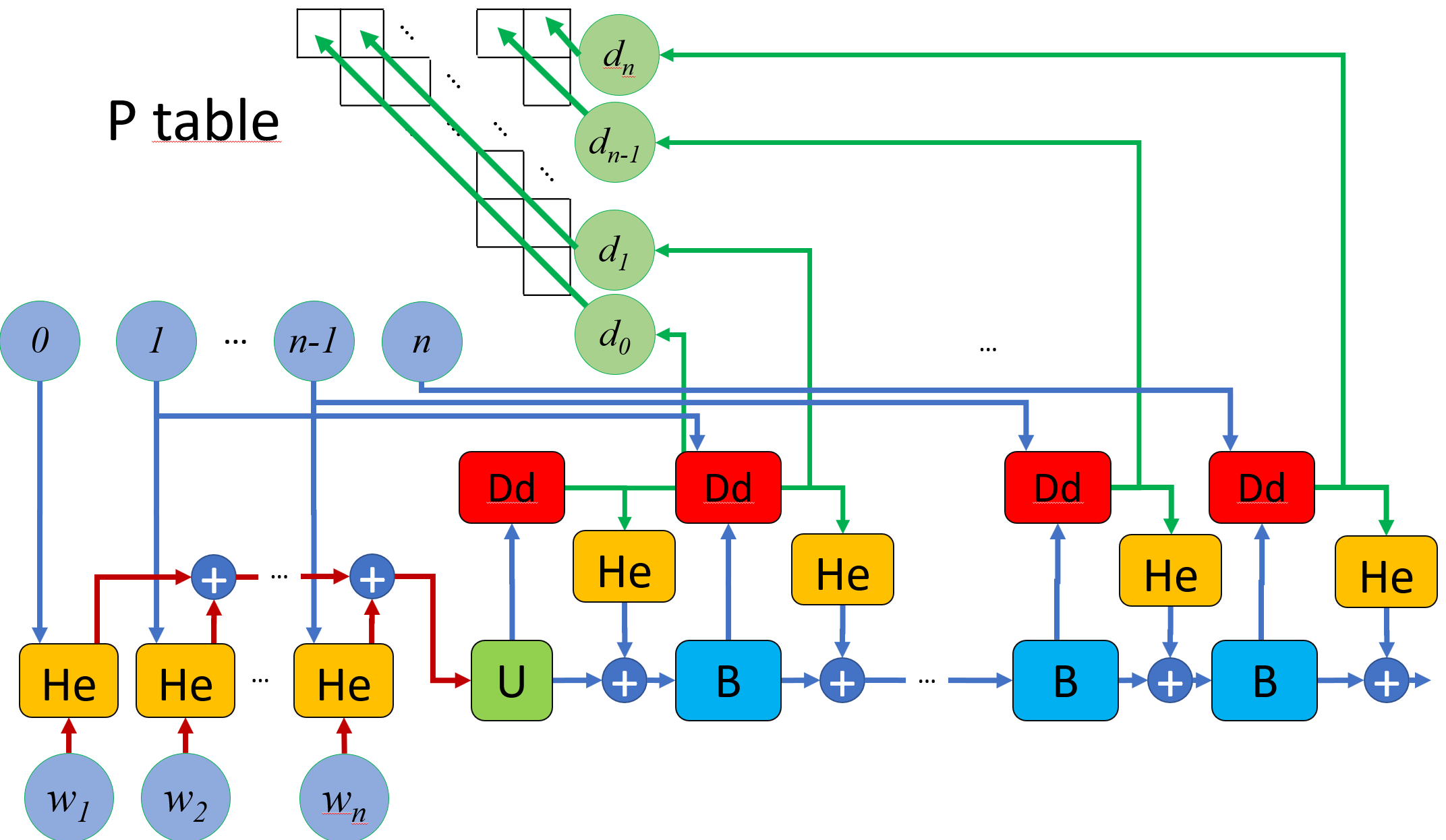}
\end{minipage}
}\hfill
\subfloat[Diagonal decoder algorithm: module \textbf{Dd}]{%
\begin{minipage}[b]{.53\linewidth}
\begin{algorithm}[H]
\setlength\baselineskip{18.5pt}
\begin{algorithmic}[1]
\setlength\baselineskip{18.5pt}
    \For {$i \gets 0$ to $n-1 - v$}
        \State $\vec{o} \gets \sigma(\CYKmatrix{W} \invsc{i} \CYKmatrix{P} \invsc{i+v+1} \vec{1})$
        \For {$s \in NT$} 
            \If {$\vec{o}_{index(s)} > t$}
                \State $\CYKmatrix{P_v} \gets \CYKmatrix{P_v} + \mysc{i}\mysc{s}\mysc{i+v+1}$
            \EndIf
        \EndFor
    \EndFor
\end{algorithmic}
\caption{\textbf{Dd}($\CYKmatrix{P}$,v) \Return ($\CYKmatrix{P_v}$,$d_v$)}
\label{alg:dd_algorithm}
\end{algorithm}
\end{minipage}
}
\caption{The CYK in a Recurrent Neural Network (CYKNN)}
\label{fig:cyknn}
\end{figure}

The inspiring idea represented by the Algorithm \ref{alg:protoCYKNN} may be translated in a more formal recurrent neural network architecture; see Fig. \ref{fig:cyknn}(a). Indeed, \emph{for} cycles in the algorithm call for a recurrent architecture or a transformer-like architecture where all the operations are done in a single pass.

The overall architecture, called CYKNN, contains then \textbf{Holographic Encoder (He)}, the \textbf{Unary Module} (\textbf{U}), the \textbf{Binary Module}, and the \textbf{Diagonal Decoder (Dd)}. These modules are hereafter defined.

The \textbf{Holographic Encoder (He)} represents the module that transforms input tokens $w_i$ into an initial matrix $\CYKmatrix{P}_0$ representing all the tokens. Each token is represented along with its span from $i$ to $i+1$, that is: 
$$
o_i = \mysc{i} \mysc{w_i} \mysc{i+1}
$$
As depicted in the architecture in Fig. \ref{fig:cyknn}, these $o_i$ are summed up as a single $\CYKmatrix{P}_0$ matrix:
$$
\CYKmatrix{P}_0 = \sum_{i=0}^{n-1} \mysc{i} \mysc{w} \mysc{i+1} 
$$

The \textbf{Unary Module} (\textbf{U}) is responsible for applying unary rules within the CYKNN architecture.
It applies the unary grammar once to obtain the first diagonal of the explicit CYK matrix in a distributed manner $P_{1}$. The \textit{holistic} operation is: 
$$
\CYKmatrix{P}_{1} = \CYKmatrix{P}_{0} \CYKmatrix{R}_{U} 
$$
that exploits the matrix $\CYKmatrix{P}_0$ and the grammar of the unitary rules encoded in $\CYKmatrix{R}_U$. The grammar is initialized with:
$$
\CYKmatrix{W}_{U} = \sum_{A \rightarrow a \in R} \invsc{a} \mysc{A}
$$
Then, for a sentence of length $n$, $R_{U}$ is obtained in the following way:
$$
\CYKmatrix{R}_{U} = \sum_{i=1}^n \invsc{i} \CYKmatrix{W}_{U} \mysc{i}
$$
After the initialization, the grammar matrix $\CYKmatrix{W}_{U}$ is trainable.

The \textbf{Binary Module} (\textbf{B}) is the module that implements the main operation of the Algorithm \ref{alg:protoCYKNN}, that is:
$$
\CYKmatrix{P}_{i+1} = \CYKmatrix{P}_{i} \CYKmatrix{R}_{B} \CYKmatrix{P}_{i}
$$
In this case, the initialization of the matrix $\CYKmatrix{W}_{B}$ representing the binary rules of the grammar is:
$$
\CYKmatrix{W}_{B} = \sum_{A \rightarrow B C  \in R} \invsc{B}\mysc{A}\invsc{C}
$$
For a sentence of length $n$, $\CYKmatrix{R}_{B}$ is obtained in the following way:
$$
\CYKmatrix{R}_{B} = \sum_{i=1}^n \invsc{i} \CYKmatrix{W}_{B} \invsc{i}
$$
As for the unary rule grammar $\CYKmatrix{W}_{U}$, also the grammar matrix $\CYKmatrix{W}_{B}$ is trainable after the initialization.

The \textbf{Diagonal Decoder (Dd)} is the most complex module as it extract cells of the diagonals from the encoded CYK matrix $\tensor{P}_{i}$. This is done for two reasons. The first reason is the construction of the final symbolic CYK matrix. The second reason is to purify the signal to be inserted in the $\tensor{P}_{i+1}$ matrix so that symbols are better presented in the novel matrix. The key component of the module is the following operation:
$$
\vec{o} = \sigma(\CYKmatrix{W}_{NT} (\invsc{i} \CYKmatrix{P} \invsc{j})_0)
$$
where $\CYKmatrix{W}_{NT}^T$ is the decoding matrix for non terminal symbols (NT), 
$\sigma(\:)$ is a sigmoid function and  $(\invsc{i} \CYKmatrix{P} \invsc{j})_0$ is the first column of the circulant matrix of $\invsc{i} \CYKmatrix{P} \invsc{j}$ that represents the sum of the representations of the symbols in the cell $(i,j)$. The first column of the circulant matrix is the vectorized version of the sum. Then, each dimension $o_i$ of the vector $\vec{o}$ represent the presence of the symbol $s \in NT$ where $index(s) = i$ in the cell. 
The \textbf{Dd} algorithm is reported in Figure \ref{fig:cyknn}(b), where the output is $d_v$ that contains the cell of the CYK table of the diagonal $v$ and $\CYKmatrix{P_V}$ representing its distributed representation version.

\subsection{Experimental investigation}

In this section, we aim to demonstrate CYKNN's ability to perform CYK parsing and we compare it with current transformer-based approaches to perform the same task.  

\subsubsection{Experimental set-up}

To experiment with the parsing ability of existing large language models and of our novel neural network architecture, CYKNN, that realizes a CYK with differentiable matrix-vector multiplications, we first set up a very simple grammar $G_0 = (T,NT,S,R_U \cup R_B)$ (see Appendix \ref{sec:grammar0}) and we parsed 5,414 randomly-generated sequences with a traditional symbolic CYK (see Appendix \ref{sec:samples} for an example of the dataset).
The dataset has been then split in the three classical parts: training with 3,000 sequences, testing with 2,000 sequences, and development with 414 sequences. 

Evaluation measures assess the ability of each method to reconstruct the oracle CYK matrix. Then, we used the classical precision, recall, and f-measure over the CYK matrices produced by the systems. Just for completeness, we here report the definition of these 3 measures on the CYK matrices $P_{oracle}$ and $P_{system}$. Defining two sets $O$ and $S$, respectively, for the oracle and the system $P$ matrices as $O = \{(i,j,Sym) | Sym \in P_{oracle}[i,j]\}$ and $S = \{(i,j,Sym) | Sym \in P_{system}[i,j]\}$, precision, recall, and f-measure are classically defined as $Prec = |O \cap S|/|S|$, $Rec = |O \cap S|/|O|$, and $f\-measure = 2Prec \cdot Rec/(Prec+Rec)$. 
For the evaluation of the larger language models, we produced a subset \emph{Small Test} of the testing set that contains only sequences of lengths 5,6,7, and 8. The complete test dataset is referred as \emph{Full Test}.

In these experiments, we aim to determine whether current transformer-based LLMs can solve this neuro-symbolic task. For this reason, we experimented with two classes of approaches: (1) a finetuning approach for relatively large language models; (2) an in-context learning approach with larger language models. For the first class, we used two models of the Qwen family: Qwen2.5-1.5B-Instruct \citep{qwen2025qwen25technicalreport} and Qwen3-8B \citep{yang2025qwen3technicalreport}. Both Qwen2.5-1.5B-Instruct and Qwen3-8B were fine-tuned with LoRA \citep{10.1145/3676151.3719377}, using 20\% of the parameters, for 3 training epochs on the Train set for training. Testing has been performed on the Full Test Set and extrapolated for the Reduced Test Set.
For the second class, we used two models, gpt-oss-20B \citep{openai2025gptoss120bgptoss20bmodel} and Gemma-3-27B-it, referred as Gemma-27B \citep{gemmateam2025gemma3technicalreport}. In this setting, no additional training was performed; instead, we designed a few-shot prompt (see Appendix \ref{sec:prompts}) to guide the models toward producing a well-formed CYK parse matrix, and evaluated them on strings of length 5, 6, 7, and 8 drawn from the same test set used for the other models.

To evaluate the ability of the CYKNN to model grammars in matrices, we experimented with 3 additional grammars, $G_1$, $G_2$, and $G_3$ built by augmenting $G_0$ with additional rules to make more complex the encoding of the grammars in the distributed matrices $W_B$ and $W_U$ representing the binary and the unary rules of the grammar itself. Grammars $G_1$, $G_2$, and $G_3$ have 20 unary rules and, respectively 5, 8, and 21 binary rules. Then, grammar $G_1$ increases only the number of unary rules with respect to grammar $G_0$. 

CYKNNs have been tested with different sizes $d$ of the embedding vectors: $100$, $1,000$, and $2,000$. We trained CYKNNs with randomly initialized grammar matrices $W_U$ and $W_B$ and with these matrices initialized to express $G_0$, $G_1$, $G_2$, and $G_3$. The version with random intialized matrices is referred to as CYKNN G=Random. Versions with initialized grammar matrices are referred to as CYKNN G=$G_i$.

Experiments with the CYKNN family have been performed on a server with a single NVIDIA GeForce GTX 1070 with 8 Gb of on-board RAM. Experiments run for 3 days. Fine-tuning of Qwen family has been performed on a server with two NVIDIA RTX A6000 with 48Gb of on-board RAM.

\begin{table}
\caption{Precision, recall, and f-measure on the reduced and the full test set for CYKNN, for two medium-sized pre-trained LLMs with fine-tuning (Qwen2.5-1.5B and Qwen3-8B), and for two large-sized LLMs with in-context learning (Gemma-27B and gpt-oss-20B)}
\label{tab:performances}
\centering
\begin{tabular}{l|c c c | c c c}
 \hline
 \hline
                  & \multicolumn{3}{c|}{Small Test} & \multicolumn{3}{c}{Full Test}  \\
\textit{Model}    & \textit{Prec} & \textit{Rec} & \textit{f-measure} & \textit{Prec} & \textit{Rec} & \textit{f-measure}  \\
 \hline
Qwen2.5-1.5B-Instruct  &               0.574  & 0.036 & 0.067 & 0.521  & 0.216 & 0.305\\ 
Qwen3-8B  &                   0.378  & 0.004 & 0.007 & 0.689  & 0.219 & 0.333\\ 
 \hline
gpt-oss-20B  &                  \textbf{0.940}  & 0.589 & 0.724 & -  & - & -\\ 
Gemma-27B  &                0.258  & 0.355 & 0.299 & -  & - & - \\ 
 \hline
CYKNN G=Random  & 0.936  & 0.264 & 0.412 & \textbf{0.900}  & 0.173 & 0.290\\ 
CYKNN without CF &             0.749  & 0.583 & 0.656 & 0.865  & 0.394 & 0.541\\ 
 \hline
CYKNN with $G_0$&              0.777  & \textbf{0.837} & \textbf{0.806} & 0.785  & \textbf{0.765} & \textbf{0.775}\\ 
 \hline
 \end{tabular}
\end{table}

\subsubsection{Results and Discussion}

CYKNN is a good model for performing CYK parsing as a symbolic task with a neural network architecture (see Table \ref{tab:performances}). Indeed, CYKNN with grammar $G_0$ outperforms all the models both in the Small and in the Full Sets.  

\begin{figure}[htbp]
\subfloat[Learning curve of CKYNNs with differentgrammars and with dimension $d=2,000$]{%
    \includegraphics[width=0.32\textwidth]{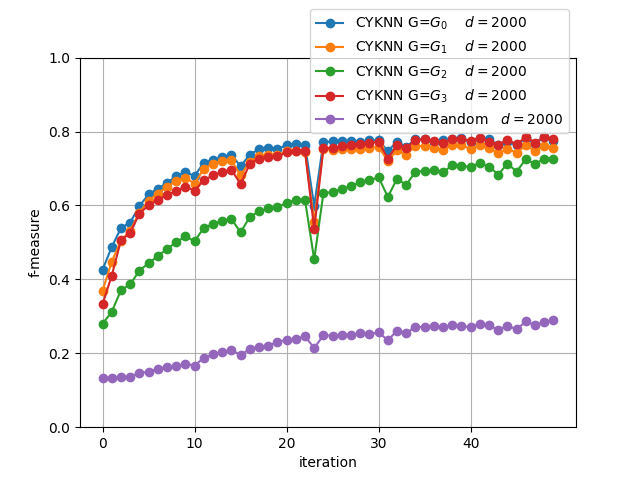}
}\hfill
\subfloat[Performance of CYKNNs with different dimensions $d$ of the representation space]{%
    \includegraphics[width=0.32\textwidth]{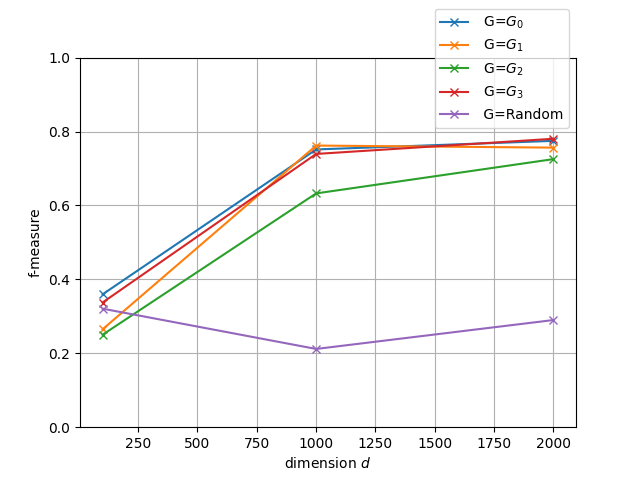}
}\hfill
\subfloat[Performance of CYKNNs with respect to the lengths of the sequences]{%
    \includegraphics[width=0.32\textwidth]{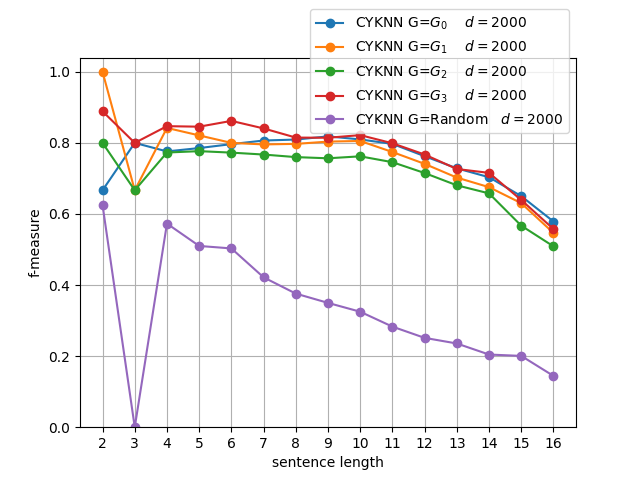}
}
\caption{Learning abilities of the CYKNN model family}
\label{fig:results_CYKNN_differnt_dims}
\end{figure}

The neuro-symbolic task of parsing with the CYK algorithm seems to be a difficult task for existing LLMs (see Table \ref{tab:performances}). The fine-tuned Qwen2.5-1.5B-Instruct and  Qwen3\_8b, and the in-context learned Gemma-27B perform poorly on the Small Set. Only the in-context learned gpt-oss-20B has higher performance in terms of f-measure. This high f-measure is mainly due to a high precision, as recall is quite low. Then, apparently,  gpt-oss-20B is conservative when filling the CYK matrix. Very strangely, performances of Qwen2.5-1.5B-Instruct and  Qwen3-8B perform are lower on shorter sequences of the Small Test than on longer sentences included in the Full Test (cf. Fig. \ref{fig:results_CYKNN_differnt_dims}(a)). 

\begin{figure}[htbp]
\subfloat[Performance over sequence lengths of two CYKNNs with dimension $d=2,000$ with grammar $G_0$ and a random grammar matrix vs. two finetuned \textit{qwen} models]{%
    \includegraphics[width=0.45\textwidth]{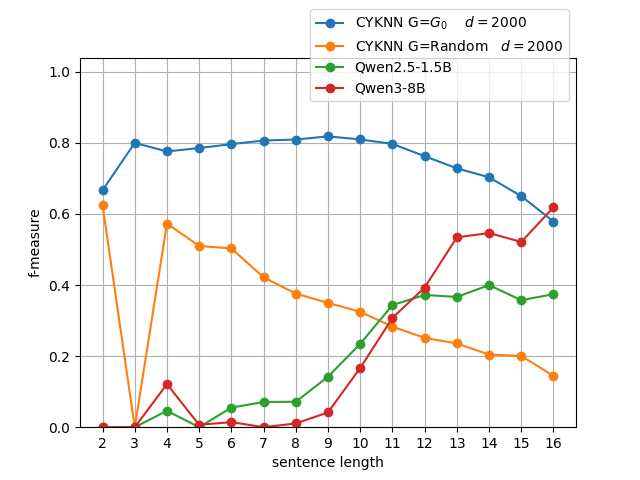}
}\hfill
\subfloat[Performance over sequences of length 5,6,8, and 9 of two CYKNNs with dimension $d=2,000$ with grammar $G_0$ and a random grammar matrix vs. two larger LLMs]{%
    \includegraphics[width=0.45\textwidth]{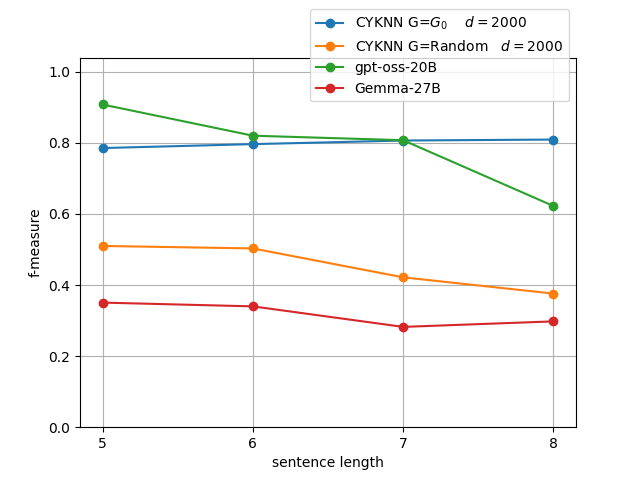}
}
\caption{CKYNNs with respect to finetuned and existing large LLMs with in-context learning: Performance in terms of f-measure with respect to the oracle on sequences of the test set.}
\label{fig:CYKNN_vs_LLMs}
\end{figure}

There also seems to be a deeper issue: the context-free, algorithmic nature of the CYK parsing task is not captured by a transformer architecture equipped with attention \citep{NIPS2017_3f5ee243}. Fine-tuning of pre-trained LLMs does not help the transformers underlying Qwen2.5-1.5B-Instruct and  Qwen3-8B to learn how to produce CYK matrices. Fine-tuned Qwen models seem to learn the association of the prefixes of sequences to the initial part of the CYK matrices. Fine-tuned Qwen models outperform CYKNN with random grammar matrix only for sequences longer than 12 tokens. This fact, along with the extremely low performance for sequences smaller than 10 tokens, suggests that the underlying transformers are somehow only memorizing and are not inducing any algorithm from data.     

Within the CYKNN family, the role of the complexity of the initialization of the grammar ($G_0$ vs. $G_1$, $G_2$, and $G_3$) is relevant with respect to the learning curves (see Fig. \ref{fig:results_CYKNN_differnt_dims}(a)). When the grammar has more distractors (e.g., $G_2$, the learning is slower. Nevertheless, the low growth of the CYKNN initialized with a random matrix shows that it is crucial to inject existing knowledge coding of the grammar into matrices. It is particularly relevant the size of the encoding dimension $d$ of the symbols (see Fig. \ref{fig:results_CYKNN_differnt_dims}(b)). Moreover, in contrast with fine-tuned LLMs, the CYKNN family behaves in a more intuitive way with respect to the length of the sequences (see Fig. \ref{fig:CYKNN_vs_LLMs}(a)), even with more complex grammars (see Fig. \ref{fig:results_CYKNN_differnt_dims}(c)). Finally, the context-free boost is extremely relevant, as the model without it (CYKNN without CF) performs worse (see Tab. \ref{tab:performances}).

\section{Conclusions}

The goal of this paper is to show that there is a radically different way to use neural networks when performing already-known algorithms: leveraging the parallelism of classical symbolic approaches and representations within neural networks to shape the neural network directly according to the algorithm. We showed the feasibility of the methodology by designing CYKNN as a neural network architecture devoted to perform the CYK parsing algorithm. CYKNN positively exploits encoded grammars and outperforms existing transformer-based LLMs that have inherent limits in performing the neuro-symbolic task of parsing with context-free grammars. Overall, results open the avenue to novel neuro-symbolic approaches where algorithms shape neural networks.     

\section{Limitations}

As this paper presents a innovative way of thinking, it has a number of limitations. First, the used grammar is extremely simple. Yet, we have shown that it is already enough complex for current transformer-based LLMs. Second, the code of the CYKNN family is not optimized, thus it leaves the impression that the computation is very slow and, consequently, the computational complexity very high. On the contrary, the computational complexity of the algorithm reduces the computational complexity of the classical CYK algorithm.

\bibliographystyle{unsrtnat}
\bibliography{my_bib}

\newpage
\appendix

\section{The simple grammar of the experiments}
\label{sec:grammar0}

Grammar $G_0 = (T,NT,S,R_U \cup R_B)$ is very simple as current transformer-based LLMs are not able to model the neuro-symbolic task of performing CYK parsing. The grammar is defined as follows:
\begin{center}
\begin{tabular}{l|l}
\hline
   $T$   &  a,b,c\\
\hline
   $NT$  &  S,A,B,C \\
\hline
   $R_U $   
&A $\rightarrow$ a\\
&B $\rightarrow$ b \\
&C $\rightarrow$ c \\
\hline
   $R_B$  
&S $\rightarrow$ A B\\
&S $\rightarrow$ B C \\
&A $\rightarrow$ B B\\
&B $\rightarrow$ C A\\
&C $\rightarrow$ C B\\
\hline

\end{tabular}
\end{center}

\section{A small sample of the dataset}
\label{sec:samples}

Samples in the dataset appear as follows:
\begin{center}
\begin{tabular}{lp{10cm}}
\textit{sequence} & \textit{Oracle CYK Matrix}\\
\hline
c c b a & (0,1,['C']) (0,2,[]) (0,3,[]) (0,4,['C']) (1,2,['C']) (1,3,['C']) (1,4,['B']) (2,3,['B']) (2,4,[]) (3,4,['A'])\\
c a c & (0,1,['C']) (0,2,['B']) (0,3,['S']) (1,2,['A']) (1,3,[]) (2,3,['C'])\\
\hline
\end{tabular}
\end{center}

\section{Properties of Holographic Reduced Representations}
\label{sec:properties}
The~starting point of a distributed representation and, hence, of HRR is how to encode symbols into vectors: symbol $a$ is encoded using a random vector $\vec{a} \in \R^d$ drawn from a multivariate normal distribution $\vec{a} \sim N(0, \CYKmatrix{I}\frac{1}{\sqrt{d}})$. These are used as basis vectors for the~Johnson--Lindenstrauss transform~\citep{Johnson1984}, as well as for random indexing~\citep{Sahlgren2005}. The~main property of these random vectors is the~following:
\begin{eqnarray*}
\vec{a}^\transpose\vec{b} & \approx & 
\begin{cases}
1 & \text{if } \vec{a} = \vec{b} \\
0 & \text{if } \vec{a} \neq \vec{b} \\
\end{cases}
\end{eqnarray*}

In this~paper, we use the~matrix representation of the~shuffled circular convolution introduced in~\citep{Zanzotto:ICML:2012}. In this~way, symbols are represented in a way where composition is just matrix multiplication and~the~inverse operation is matrix transposition. Given the~above symbol encoding, we can define a basic operation $\mysc{\:}$ and~its approximate inverse $\invsc{\:}$. These operations take as input a symbol
and~provide a matrix in $\R^{d\times d}$ and~are the~basis for our encoding and~
decoding. The~first operation is defined as:
\begin{eqnarray*}
\mysc{a} & = & \cm{A} \Phi \; ,
\end{eqnarray*}
where $\Phi$ is a permutation matrix to obtain the~shuffling~\citep{Zanzotto:ICML:2012} and~$\cm{A}$ is the~circulant matrix of the~vector $\vec{a}^\transpose = \begin{pmatrix} a_0 & a_1 & \ldots & a_{d-1}\end{pmatrix}$.
, that is: \\
$$
\cm{A} = 
\begin{bmatrix}
a_0 & a_{d-1} & \ldots & a_1\\
a_1 & a_0 & \ldots & a_2 \\
\vdots & \ddots & \vdots \\
a_{d-2} & a_{d-3} & \ldots & a_{d-1}\\
a_{d-1} & a_{d-2} & \ldots & a_0 \\
\end{bmatrix}
= 
\begin{bmatrix}
\vline &\vline &\ldots & \vline \\
s_0(\vec{a}) &s_1(\vec{a}) &\ldots & s_{d-1}(\vec{a}) \\
\vline &\vline &\ldots & \vline& \\
\end{bmatrix}
$$ 
while~$s_i(\vec{a})$ is the~circular shifting of $i$ positions of the~vector $\vec{a}$.
Circulant matrices are used to describe circular convolution. In fact, $\vec{a} \cconv{} \vec{b} = \cm{A}\vec{b} = \cm{B} \vec{a}$ where $\cconv{}$ is circular convolution.
This~operation has a nice approximated inverse in: 
\begin{eqnarray*}
\invsc{a} & = & \Phi^\transpose \cmt{A} \; . 
\end{eqnarray*}

We then have:
\begin{eqnarray*}
\mysc{a} \invsc{b} & \approx & \begin{cases}
\CYKmatrix{I} & \text{if } \vec{a} = \vec{b} \\
\CYKmatrix{0} & \text{if } \vec{a} \neq \vec{b} \\
\end{cases} 
\end{eqnarray*}
since $\Phi$ is a permutation matrix and~therefore $\Phi \Phi^\transpose=I$ and~since:
\begin{eqnarray*}
\cmt{A}\cm{B} & \approx & 
\begin{cases}
\CYKmatrix{I} & \text{if } \vec{a} = \vec{b} \\
\CYKmatrix{0} & \text{if } \vec{a} \neq \vec{b} \\
\end{cases}
\end{eqnarray*}
due to the~fact that $\cm{A}$ and~$\cm{B}$ are circulant matrices based on random vectors $\vec{a},\vec{b} \sim N(0, \CYKmatrix{I}\frac{1}{\sqrt{d}})$;~hence, $ s_i(\vec{a})^\transpose s_j(\vec{b}) \approx 1$ if both $i=j$ and~$\vec{a}=\vec{b}$, and~$s_i(\vec{a})^\transpose s_j(\vec{b}) \approx 0$ otherwise.
Finally, the~permutation matrix $\Phi$ is used to enforce non-commutativity in the matrix product such as $\mysc{a} \mysc{b} \mysc{c}$.

\section{Experiments with LLMs as CYK table generators}
\label{sec:prompts}

\subsection{Few-shot prompting for CYK table generation}

We adopt a standard few-shot prompting setup in which the model is given a fixed CNF grammar together with two input strings and their gold CYK tables, and is then asked to generate the CYK table for a new string in the same format. The prompt is reported below.

{\scriptsize
\begin{verbatim}
Task: Given a context-free grammar in Chomsky Normal Form (CNF) and an input string w,
compute the CYK table following the format of the examples and return it.

Grammar:
S -> A B
S -> B C
A -> B B
B -> C A
C -> C B
A -> a
B -> b
C -> c

Below some example of the matrix I want you to return. Follow the exact same format.

Example_1

String_1: "cacbbcca"
CYK Matrix : (0,1,['C']) (0,2,['B']) (0,3,['S']) (0,4,['S']) (0,5,['S','A']) (0,6,[]) (0,7,[]) (0,8,[])
                 (1,2,['A']) (1,3,[]) (1,4,[]) (1,5,['S']) (1,6,[]) (1,7,[]) (1,8,[])
                         (2,3,['C']) (2,4,['C']) (2,5,['B','C']) (2,6,['S']) (2,7,[]) (2,8,['S'])
                                 (3,4,['B']) (3,5,['A']) (3,6,[]) (3,7,[]) (3,8,[])
                                             (4,5,['B']) (4,6,['S']) (4,7,[]) (4,8,['S'])
                                                         (5,6,['C']) (5,7,[]) (5,8,['C'])
                                                                     (6,7,['C']) (6,8,['B'])
                                                                                 (7,8,['A'])

Example_2

String_2: "cababcac"
CYK Matrix: (0,1,['C']) (0,2,['B']) (0,3,['A']) (0,4,[]) (0,5,[]) (0,6,[]) (0,7,[]) (0,8,[])
                 (1,2,['A']) (1,3,['S']) (1,4,[]) (1,5,[]) (1,6,[]) (1,7,[]) (1,8,[])
                             (2,3,['B']) (2,4,[]) (2,5,[]) (2,6,[]) (2,7,[]) (2,8,[])
                                         (3,4,['A']) (3,5,['S']) (3,6,[]) (3,7,[]) (3,8,[])
                                                     (4,5,['B']) (4,6,['S']) (4,7,['A']) (4,8,[])
                                                                 (5,6,['C']) (5,7,['B']) (5,8,['S'])
                                                                             (6,7,['A']) (6,8,[])
                                                                                         (7,8,['C'])

Now compute:

String_3: "{current_string}"
CYK Matrix :
\end{verbatim}
}

\subsection{Instruction tuning on CYK supervision}

To further evaluate whether explicit supervision improves algorithmic behaviour, we fine-tune instruction-following models on a CYK training set. We preprocess each example into an instruction--input--output format, where the \texttt{instruction} specifies the fixed CNF grammar, the \texttt{input} contains a query asking for the CYK matrix of a given string, and the \texttt{output} consists of the corresponding gold CYK table serialized in the target linearized format. An example is shown below:

{\scriptsize
\begin{verbatim}
{
  'instruction': 'Compute the CYK matrix for the grammar:
                  S -> A B
                  S -> B C 
                  A -> B B
                  B -> C A
                  C -> C B
                  A -> a 
                  B -> b
                  C -> c,
  
  'input': 'Tell me the CYK matrix of cacbaccbcba',
  
  'output': "(0,1,['C']) (0,2,['B']) (0,3,['S']) (0,4,['S']) (0,5,['A']) (0,6,[]) (0,7,[]) (0,8,[]) (0,9,[]) (0,10,[]) (0,11,[])
                          (1,2,['A']) (1,3,[]) (1,4,[]) (1,5,['S']) (1,6,[]) (1,7,[]) (1,8,[]) (1,9,[]) (1,10,[]) (1,11,[])
                          (2,3,['C']) (2,4,['C']) (2,5,['B']) (2,6,['S']) (2,7,[]) (2,8,[]) (2,9,[]) (2,10,[]) (2,11,['S'])
                          (3,4,['B']) (3,5,[]) (3,6,[]) (3,7,[]) (3,8,[]) (3,9,[]) (3,10,[]) (3,11,[])
                          (4,5,['A']) (4,6,[]) (4,7,[]) (4,8,[]) (4,9,[]) (4,10,[]) (4,11,[])
                          (5,6,['C']) (5,7,[]) (5,8,[]) (5,9,[]) (5,10,[]) (5,11,['C'])
                          (6,7,['C']) (6,8,['C']) (6,9,[]) (6,10,[]) (6,11,['B','C'])
                          (7,8,['B']) (7,9,['S']) (7,10,['S']) (7,11,['A'])
                          (8,9,['C']) (8,10,['C']) (8,11,['B'])
                          (9,10,['B']) (9,11,[])
                          (10,11,['A'])"
}
\end{verbatim}
}

These serialized instruction examples are then used to fine-tune medium-sized instruct models with LoRA, allowing the models to adapt to the CYK table-generation task while keeping the number of trainable parameters limited. At test time, evaluation is performed in a zero-shot setting, where the fine-tuned model receives only the \texttt{instruction} and the \texttt{input}, and is required to generate the CYK matrix without any additional demonstrations.

\newpage

\end{document}